\documentclass[final]{article}

\usepackage{nips_2016}
\usepackage{graphicx}
\graphicspath{ {/} }

\usepackage[utf8]{inputenc} 
\usepackage[T1]{fontenc}    
\usepackage{hyperref}       
\usepackage{url}            
\usepackage{booktabs}       
\usepackage{amsfonts}       
\usepackage{nicefrac}       
\usepackage{microtype}      
\usepackage{amsmath,amssymb}

\usepackage{caption}
\usepackage{subcaption}

\DeclareMathOperator*{\E}{\mathbb{E}}
\DeclareMathOperator{\prob}{\mathbb{P}}

\title{Face Super-Resolution Through Wasserstein GANs }

\author{
Zhimin Chen, Yuguang Tong\\
University of California, Berkeley \\
  \texttt{\{mandy\_chen, ygtong\}@berkeley.edu} \\
}

\begin{document}

\maketitle

\begin{abstract}
Generative adversarial networks (GANs) have received a tremendous amount of attention in the past few years, and have inspired applications addressing a wide range of problems. Despite its great potential, GANs are difficult to train. Recently, a series of papers (Arjovsky \& Bottou, 2017a; Arjovsky et al. 2017b; and Gulrajani et al. 2017) proposed using Wasserstein distance as the training objective and promised easy, stable GAN training across architectures with minimal hyperparameter tuning. In this paper, we compare the performance of Wasserstein distance with other training objectives on a variety of GAN architectures in the context of single image super resolution. Our results agree that Wasserstein GAN with gradient penalty (WGAN-GP) provides stable and converging GAN training and that Wasserstein distance is an effective metric to gauge training progress. 
\end{abstract}

\section{Introduction}
We generally prefer images with fine details. Image super-resolution is important and it is widely used in applications ranging from photographic post-processing, medical imaging, to surveillance and security. Here we propose to use generative adversarial networks (GANs) to increase the resolution of a down-scaled face image.

GANs have garnered significant interest lately, along with incremental improvements on the framework and its numerous applications. GANs excel as generative models that represent and manipulate high-dimensional probability distributions. GANs are powerful because they transform a generative modeling problem into a zero-sum game between two adversarial networks: the generator network produces artificial data from noise, while the discriminator network is trained to discern between true data and the generator's synthetic output (Goodfellow et al., 2014).  In the past few years, the deep learning community has witnessed creative application of GANs in various tasks, including image and text-to-image translation (Isola et al., 2016; Reed et al., 2016; Zhang et al., 2016), interactive art creation (Zhu et al., 2016) and reinforcement learning (Finn et al., 2016).

Despite its great potential and popularity, the original version of GAN is affected by several issues. First, training GANs is hard; one needs to carefully design and balance the training of the generator and the discriminator and the network architecture for training to converge. Second, the canonical loss function does not correlate with output sample quality, rendering it hard to decide when to stop training or perform hyperparameter tuning. Third, generated samples may suffer from a lack of diversity in output, or "mode collapse". Recently, a series of papers (Arjovsky \& Bottou, 2017a; Arjovsky et al., 2017b; and Gulrajani et al., 2017) proposed remedies for the aforementioned issues and provided rigorous mathematical foundations to their claims. In particular, Arjovsky \& Bottou (2017a) proposed an alternative loss function, the Wasserstein distance, with theoretical properties that address shortcomings of GANs. Wasserstein-GAN (WGAN) replaces the original Jensen-Shannon divergence-based loss functions with the Wasserstein distance. However, this change gives rise to a new problem wherein the weights of the discriminator must lie within a certain range (the space of 1-Lipschitz functions), which can be solved by weight clipping. Gulrajani and colleages (2017) observed that weight clipping compromised the stability and performance of WGAN, and they proposed an improved model of WGAN with gradient penalty in place of weight clipping.

We compare the performance and efficacy of three versions of GANs in the context of face super-resolution: the original GAN, WGAN and the improved WGAN. We relax the architecture of the three models and evaluated the stability of training and the quality of generated images. We hope to assess whether the proposed modifications to GANs address its limitations.

\subsection{Contributions}
\begin{itemize}
    \item To our knowledge, we instigate the first systematic investigation on the application of the newly proposed Wasserstein GAN objective (and its improved version) on the single image super-resolution problem.
    \item We check the claims of WGAN, namely stable and converging training on a variety of architectures, and the effectiveness of the Wasserstein distance as an indicator of training progress. 
 
\end{itemize}
\section{Data}
 We use the aligned and cropped CelebA dataset (Liu et al. 2015), which contains 202,599 face images from 10,177 identities. Data preprocessing involves cropping images in a random box of $128\times 128$ pixels and resizing to $64\times 64$ pixels. These square $64\times 64$-pixel images are treated as training labels.

We aim to train state-of-the-art GANs that enhance pixel resolution along each input dimension by 4 times. Input images are $16\times 16$ pixels and are obtained by downsampling the $64\times 64$-pixel images.

\section{Method}
We implement and train GANs to generate $64\times 64$ face images from $16\times 16$ input. We examine combinations of different training objectives using different GAN architectures to evaluate the performance of WGAN and WGAN-GP objectives, which are introduced in the following subsections.

\subsection{Generative Adversatrial Network}
In a seminal paper, Goodfellow et al. (2014) introduced GAN as a powerful generative model for a wide range of applications. The basic idea of GAN is to set up a game between two players (often neural network), a generator and a discriminator. The discriminator uses canonical supervised learning to perform classification. The generator produces samples (also referred to as fake data) from input, usually low dimensional noise, and passes them to the discriminator to determine their similarity to real data. Therefore the generator tries to fool the discriminator while the discriminator tries to distinguish fake data from real data. A properly-designed game should result in the fake distribution resembling the real distribution. 

The game between discriminator D and generator G is most often formulated as a zero-sum/ min-max game with cross entropy objectives (Goodfellow et al., 2014)

\begin{equation}
\min_G\max_D \E_{x\sim \prob_r}[\log(D(x))] + \E_{\tilde{x}\sim \prob_g}[\log(1 - D(\tilde{x}))]
\end{equation}
where $\prob_r$ is the real data distribution, and $\prob_g$ is the fake/generated distribution. The fake distribution is implicitly defined by the samples produced by the generator  $\tilde{x}\sim G(z)$, where $z\sim p(z)$ comes from some distribution $p$. Hence the discriminator loss is 
\begin{equation}
J_D = -\E_{x\sim \prob_r}[\log(D(x))] - \E_{\tilde{x}\sim \prob_g}[\log(1 - D(\tilde{x}))]
\label{dcgan_dloss}
\end{equation}
while the generator loss is simply the negation:
\begin{equation}
J_G = - J_D
\end{equation}
Goodfellow et al. (2014) showed that learning in this min-max game is equivalent to minimizing the Jensen-Shannon (JS) divergence between the real distribution and fake distribution.

In practice, a modified generator loss is used
\begin{equation}
J_G = -\E_{\tilde{x}\sim \prob_g}[\log(D(\tilde{x}))]
\label{dcgan_gloss}
\end{equation}
We referred to the loss defined in Eq. (\ref{dcgan_dloss}) and Eq. (\ref{dcgan_gloss}) as the \textbf{GAN} objective. It has been often observed that training GANs with the above objective is unstable and non-convergent, in that the generator and the discriminator repeatedly undo each others' progress (Goodfellow, 2016). Several notable phenomena are:
\begin{enumerate}
    \item Mode collapse, when several different input to generator lead to the same output. In the simple example of learning MNIST data, a generator generating predominantly 1s and very few 2s and 5s would demonstrate mode collapse. In other words, a generator suffering from mode collapse problems would have difficulty producing diversified samples.
    \item Weak correlation between training loss and training efficacy. Training GANs over longer periods may not yield a better generator. This is fatal because we do not know when we should stop training. 
    \item Oscillation in training. The Generator alternates between generating two kinds of samples, without converging.
\end{enumerate}

Very recently, a series of papers (Arjovsky et al., 2017a, Arjovsky et al., 2017b, Gulrajani et al., 2017) analyzed the reason for the above failures and proposed solutions that claim to make progress in solving the above issues by reconstructing training objectives. 

\subsection{Wasserstein GAN (WGAN)}
\label{section:wgan}
Arjovsky and collaborators (2017a, 2017b) argued that the JS divergence, along with other common distances and divergences, are potentially not continuous and thus do not provide a usable gradient for the generator. The authors proposed using Wasserstein distance $W (f, g)$ to measure the difference between two distributions. The Wasserstein distance is informally defined as the minimum cost of transporting mass in order to transform the distribution $f$ into the distribution $g$. The Wasserstein distance has the desirable property of being continuous and differentiable almost everywhere under mild assumptions. 
\begin{equation}
\min_G\max_{D\in L} \E_{x\sim \prob_r}[D(x)] - \E_{\tilde{x}\sim \prob_g}[D(\tilde{x})]
\end{equation}
where $L$ is the set of 1-Lipschitz functions. To implement Wasserstein GAN, Arjovsky et al. (2017b) suggested the following modifications to GAN:
\begin{enumerate}
    \item Remove sigmoid or softmax activation in the output layer. 
    \item Discriminator loss $J_D=\E_{\tilde{x}\sim \prob_g}[D(\tilde{x})]-\E_{x\sim \prob_r}[D(x)]$.
    \item Generator loss $J_G = - \E_{\tilde{x}\sim \prob_g}[D(\tilde{x})]$.
    \item Confine discriminator weights $W_D$ to range (-c, c), i.e. $W_D\leftarrow\texttt{clip}(W_D, -c, c)$.
    \item Avoid using momentum based optimizer such as Adam. Use RMSProp or SGD.
\end{enumerate}
The authors introduced clipping to force discriminator to be 1-Lipschitz. The choice of optimizer is empirical and was not justified.

\subsection{Wasserstein GAN with Gradient Penalty (WGAN-GP)}
WGAN has attracted much attention since its proposal and has been applied in many scenarios. Gulrajaniet et al. (2017) suggested that WGAN can still generate low-quality samples in some settings, despite significant progress towards stable GAN training. These authors analyzed the weight distribution and argued that the weight clipping in WGAN lead to pathological behavior. They proposed penalizing the norm of the gradient of the discriminator with respect to its input as an alternative method for enforcing the Lipschitz constraint. The paper claimed that their modified WGAN-GP objective stabilizes GAN training over a wide range of architectures with almost no hyperparameter tuning.

Specifically, WGAN-GP removes weight clipping from WGAN and adds gradient penalty to discriminator loss:
\begin{equation}
J_D=\E_{\tilde{x}\sim \prob_g}[D(\tilde{x})]-\E_{x\sim \prob_r}[D(x)] + \lambda \E_{\hat{x}\sim \prob_{\hat{x}}}[\| \nabla_{\hat{x}}D(\hat{x})\|_2 - 1)^2]
\end{equation}

\subsubsection{Perceptual Loss}
\label{section:perceptual-loss}
To ensure that the generated images resemble the input image, we require the generator not only minimize the adversarial loss ($J_G$ in Section \ref{section:wgan}) but also the $L_1$ difference between the input image and the downsized generated image. Hence we define the new generator loss function to be a weighted sum of the above two terms
\begin{equation}
J_G^{tot} = (1-\gamma) J_G + \gamma \E_{\tilde{x}=G(z)\sim \prob_g}\| f_d(\tilde{x})- z\|_1
\end{equation}
where $f_d$ is a downscale function that scles a $64\times64$-pixel image to $16\times 16$ pixels.

\section{Architectures}
\label{section:architectures}
We apply three different training objectives, namely, GAN, WGAN and WGAN-GP on several architectures: 
\begin{enumerate}
    \item DCGAN: D is a convolutional neural network (CNN) with 4 fully-convolutional layers and a single full-connected layer for output. G is a deconvolutional network with 2 convolutional layers and 4 deconvolutional layers.
    \item ResNet: D is the same CNN as DCGAN. G is a residual network (ResNet) with 3 residual blocks and 3 fully-convolutional layers. 
    \item MLP: D and G are both multi-layer perceptrons (MLP) with 4 fully connected layers.
\end{enumerate}
By default, all three architectures perform batch-normalization and use ReLU activations. 
 In some additional runs, we remove batch-normalization and replace ReLU with nonlinear $\tanh$ activation. All runs with the WGAN objective use the RMSProp optimizer, following the original WGAN paper (Arjovsky et al., 2017b). All runs with GAN or WGAN-GP objective use the Adam optimizer. Optimizer parameters follow the recommendations in Gulrajani et al. (2017):
\begin{itemize}
\item GAN (Adam, $\alpha$ = .0002, $\beta_1$ = 0.5)
\item WGAN with weight clipping (RMSProp, $\alpha$ = .0001)
\item WGAN-GP with gradient penalty(Adam, $\alpha$ = .0002, $\beta_1$ = 0.5, $\beta_2$ = 0.9)
\end{itemize}
For all runs unless specified otherwise, the weight of $L_1$ loss uses a fixed value $\gamma = 0.9$. For runs with WGAN-GP objective, $\lambda$, the weight of gradient penalty, is set to 10 (Gulrajani et al., 2017).

The source code and instructions to replicate this work are publicly available on GitHub \footnote{\href{https://github.com/MandyZChen/srez}{https://github.com/MandyZChen/srez}}
\footnote{\href{https://github.com/YuguangTong/improved\_wgan\_training}{https://github.com/YuguangTong/improved\_wgan\_training}}.

\section{Results and Discussions}
We focus on three claims about WGAN in Arjovsky et al. (2017b) and Gulrajani et al. (2017)
\begin{enumerate}
    \item Training quality correlates strongly with sample quality. Longer training leads to smaller discriminator loss (absolute value) and better generated samples.
    \item Training is stable across multiple architectures with minimum hyperparameter tuning.
    \item Training converges progressively to a good local minimum, without suffering common non-convergence problems such as mode collapse.
\end{enumerate}

\begin{figure}[ht]
\begin{center}
  \includegraphics[scale=0.5]{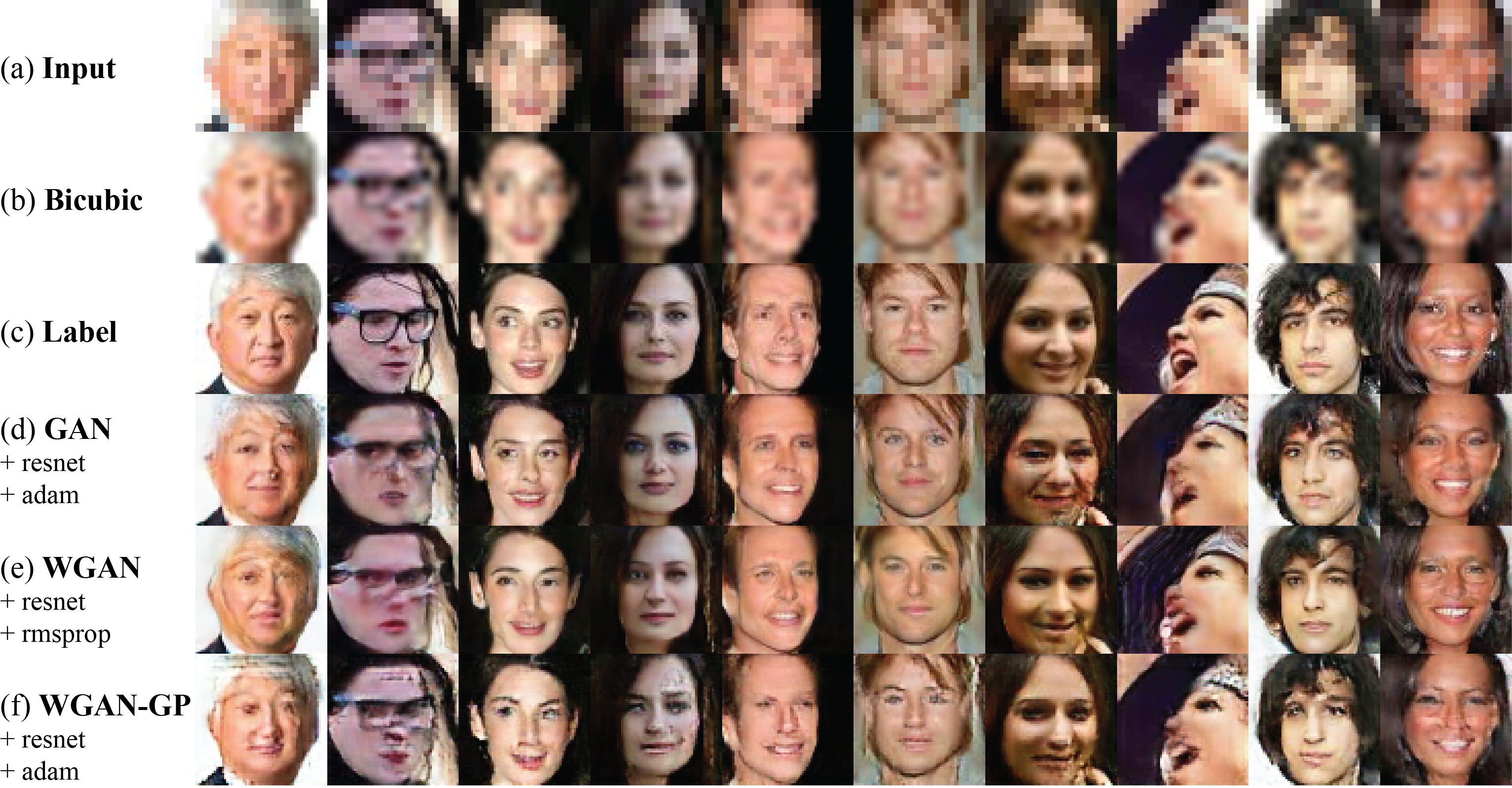}
\end{center}   
\caption{(a) Input of 4 timed down-scaled images. (b) Bicubic interpolation of the input images. (c) True labels. (d-f) three variations of GANs trained with recommended set of optimizer and hyperparameters, and ResNet (deep residuals networks).}
\label{fig:resnet-output}
\end{figure}

\begin{figure}[hb]
\begin{center}
    \includegraphics[width=\textwidth]{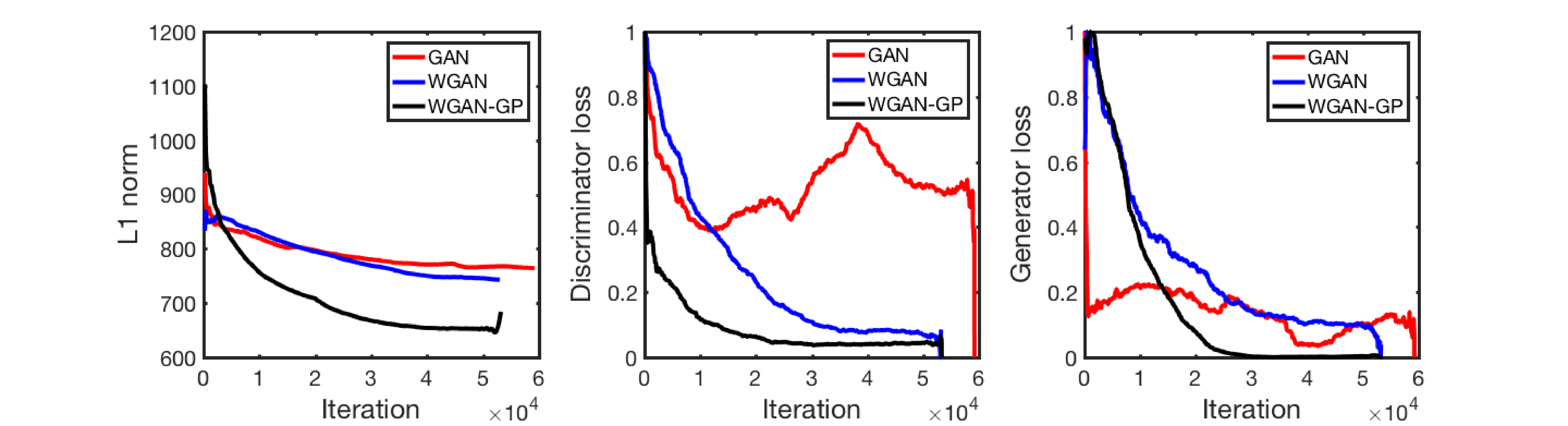}
\end{center}
\caption{Plots of $L_1$ loss (left), discriminator loss (middle) and generator loss (right) over generator iterations for three models: the original form of GAN with Adam, WGAN with weight clipping and RMSprop, WGAN with gradient penalty and Adam. All training curves were passed through the same moving average filter with a window of 100 iterations.}
\label{fig:training-curves}
\end{figure}

\subsection{Sample quality and training curves}
Figure \ref{fig:resnet-output} shows generator outputs from three GANs with the same ResNet architecture (described in Section \ref{section:architectures}) but different loss functions or training objectives. Panel (a) shows the test input which are $16\times16$-pixel images, (b) $64\times64$-pixel images obtained from (a) by bicubic interpolation, (c) the original images (label). Panel (d)-(f) demonstrate outputs from GANs with the GAN objective, WGAN objective and WGAN-GP objective respectively. Overall, all three GANs produce images of satisfying quality that significantly outperform the baseline images obtained by bicubic interpolation. The side profile faces (image column 2 and 8) look less reasonable than frontal faces. This is expected because the majority of the training images are frontal faces. We notice that WGAN-GP generates better glasses (2nd sub-figure in each panel), which is also a difficult feature.

Figure \ref{fig:training-curves} presents three training curves. The left panel shows the $L_1$ loss of the generated imaged images (defined in Section \ref{section:perceptual-loss}). This metric drops monotonically for all three GANs, showing that training does progress. It is worth pointing out that $L_1$ loss is not a perfect metric of image quality, since naive interpolations would probably have very small error. In fact, several images in (f) looks noisier than (d) and (e), yet (f) has lower $L_1$ loss. 

The middle and right panels show the the discriminator loss and generator loss versus training steps. Using WGAN and WGAN-GP objectives, the discriminator loss decreases monotonically, correlating perfectly with training progress, consistent with the claim in Arjovsky et al. (2017b) and Gulrajani et al. (2017). In contrast, the widely used GAN objective leads to oscillating discriminator loss, and therefore does not effectively reflect training progress. We conclude that the WGAN and WGAN-GP objectives honor promises to correlate well with training progress, which is vital for hyperparameter tuning, detecting overfitting or simply deciding when to terminate training.

\subsection{Architecture robustness}
\begin{figure}[ht]
  \begin{center}
    \includegraphics[scale=0.5]{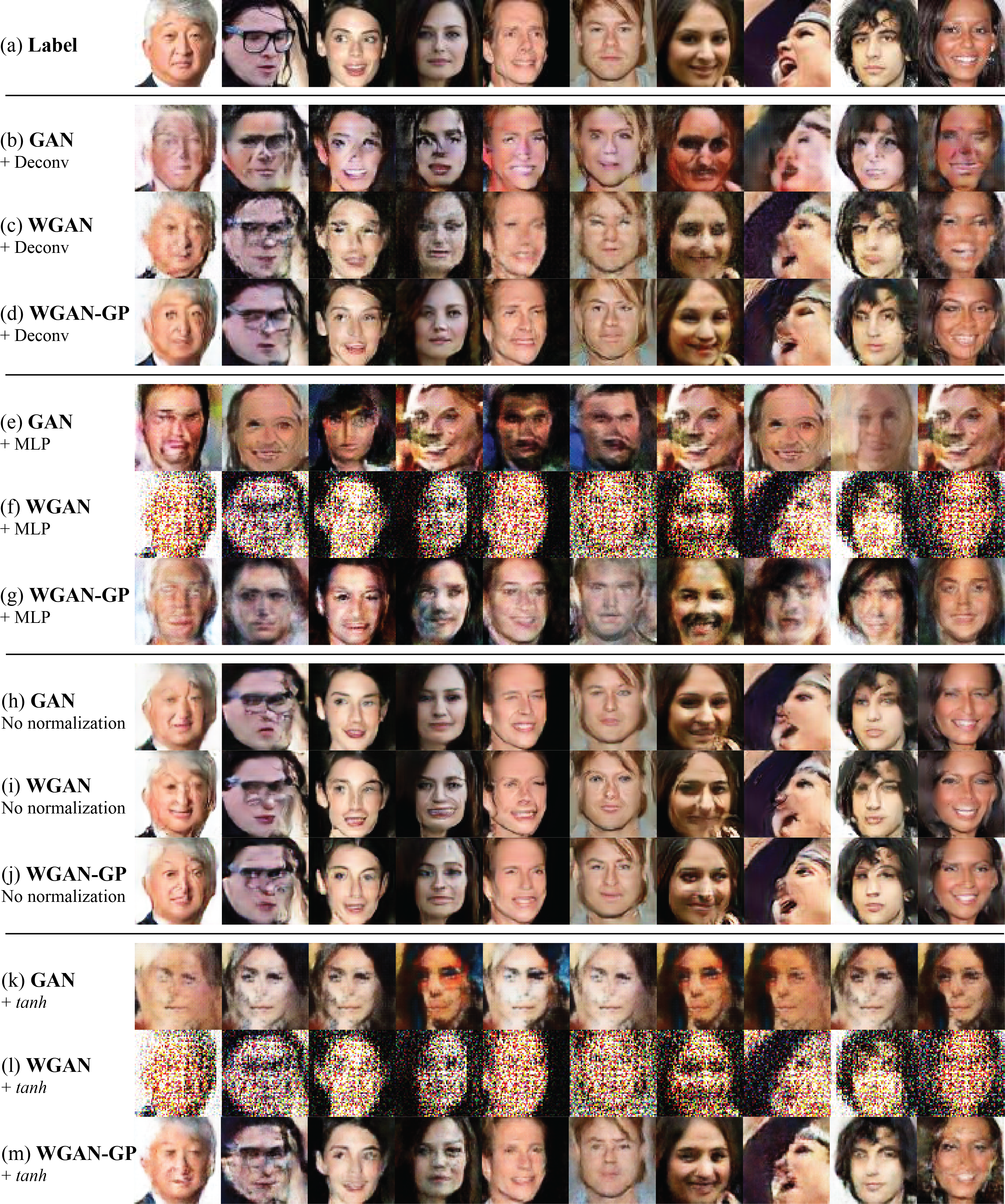}
\end{center}
\caption{Output samples of GANs trained with different architecture.}
\label{fig:other-output}
\end{figure}

\begin{figure}
    \centering
    \includegraphics[scale=0.5]{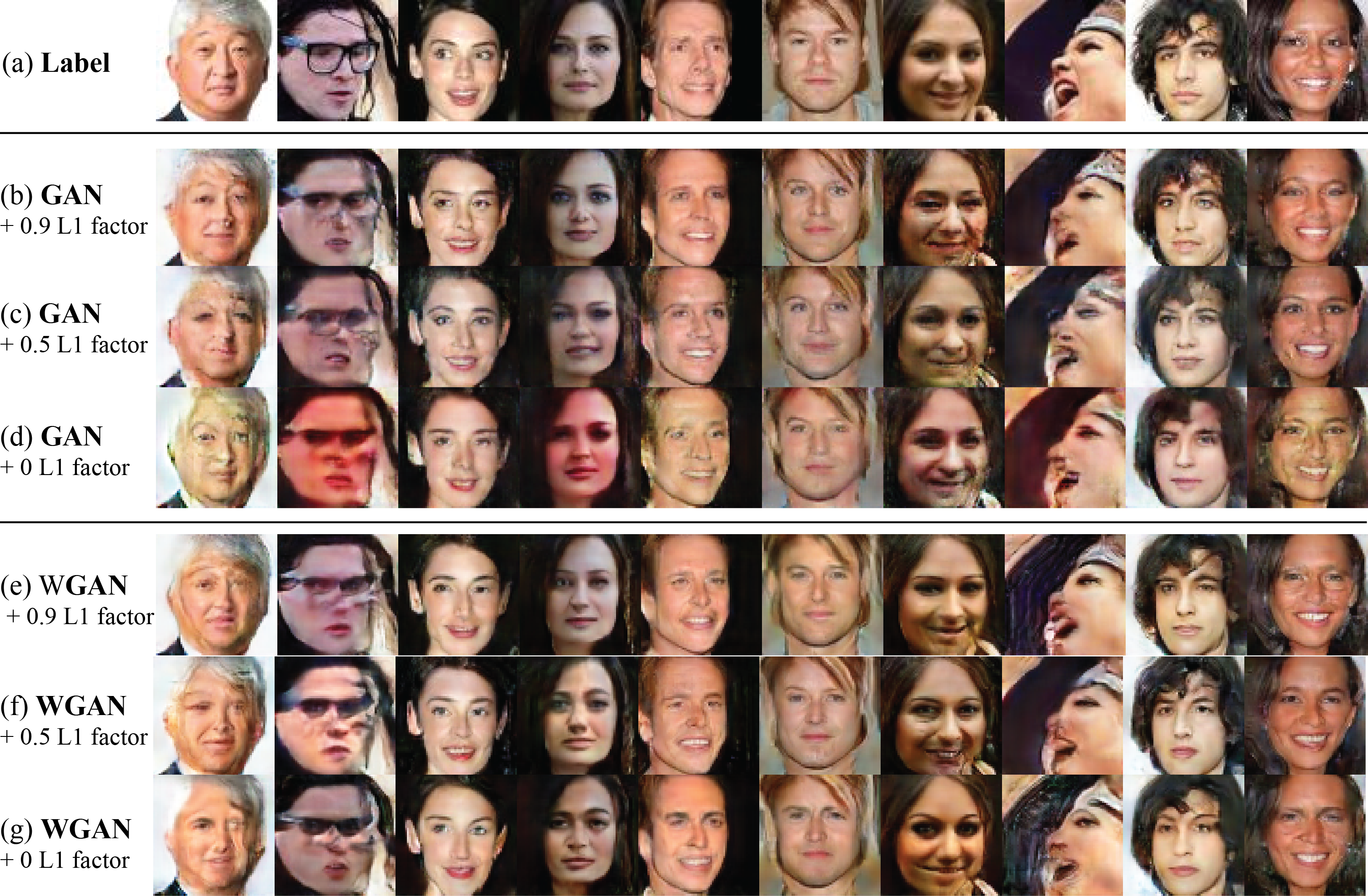}
    \caption{Models trained with different weight of $L_1$ loss: $\gamma = 0, 0.5, 0.9$. }
    \label{fig:L1-weight}
\end{figure}

Figure \ref{fig:other-output} show generated images from GANs using different architectures and training objectives. Panel (b)-(d) show results from DCGANs. WGAN and WGAN objectives seem to produce sharper images. Panel (e)-(g) show images generated by GANS using MLP architecture. While WGAN-GP is still able to produce modest results, GAN and WGAN objectives fail for this simple architecture. Panel (h)-(j) show generated images from ResNet GANs without batch normalization. It turns out batch normalization is not necessary in ResNet architecture. Panel (k)-(j) experiment with nonlinear activation function on DCGANs. Again GAN and WGAN objectives fail completely whereas WGAN-GP perform reasonably well. Overall, our experiments support the claim that WGAN-GP enables stable training without stringent constraint on architecture design. 

Figure \ref{fig:L1-weight} demonstrates the effect of $\gamma$ parameter, the weight of $L_1$ loss term in the generator loss (see Section \ref{section:perceptual-loss}). The architecture is fixed to be ResNet and the training objective is either GAN or WGAN. We expect that larger $\gamma$ would bias generators to produce images more similar to input images with the possible risk of reducing perceptual resolution. Panel (d) shows that with the GAN objective, setting $\gamma=0$ causes noticeable color shift in the output samples. Panel (e)-(g) seem to suggest that the WGAN objective leads to more robust convergence over different values of $\gamma$. Notice we would expect the images in panel (d) and (g) to be more dissimilar to the label images since no $L_1$ loss exists to force generated images to be similar to input. The ResNet design may have effectively imposed a penalty for diverging from input images.

\subsection{Mode collapse}
\begin{figure}
	\begin{subfigure}[t]{\textwidth}
    \centering
        \includegraphics[width=0.72\textwidth]{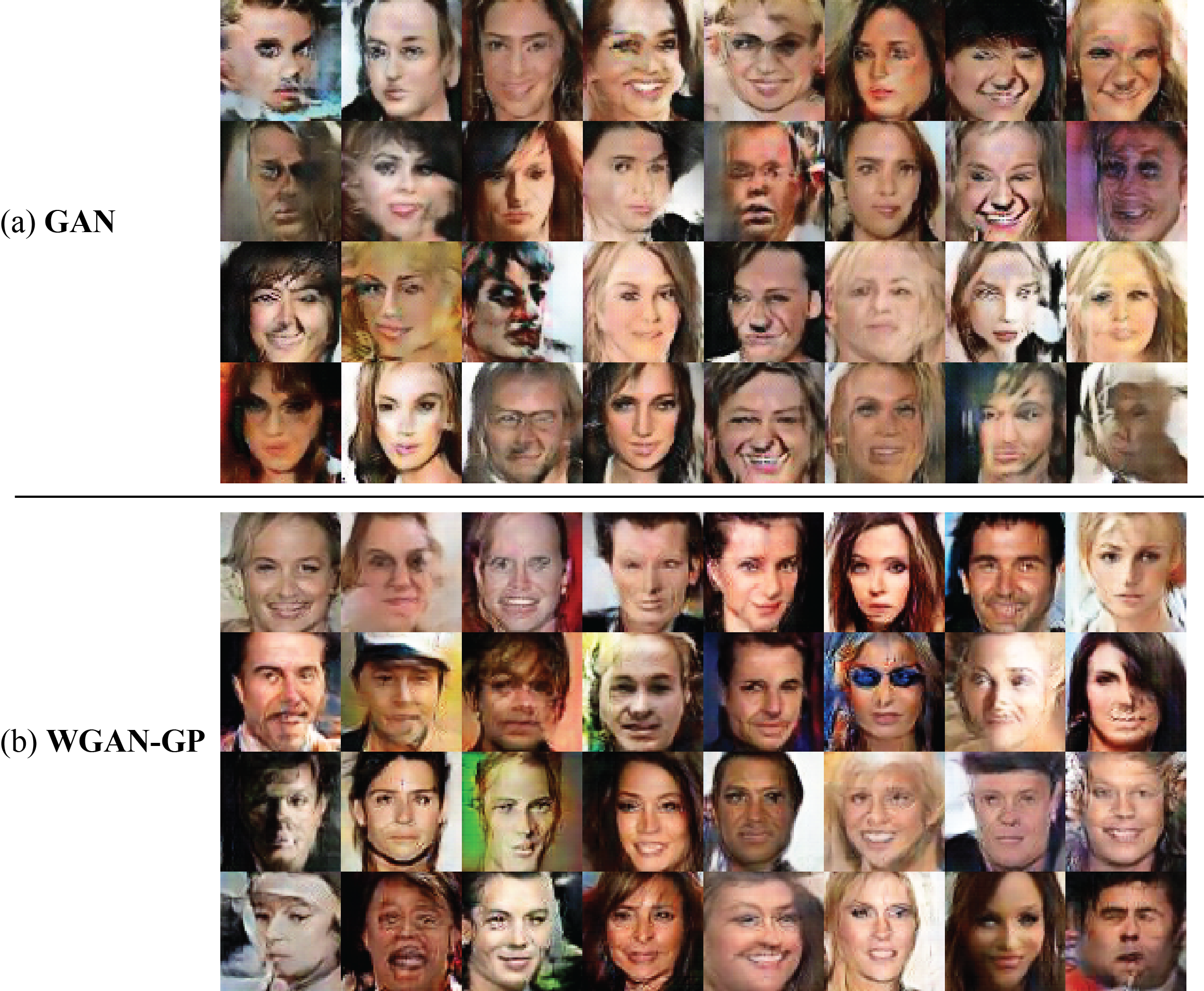}
    \end{subfigure}
    \caption{Non-cherry-picked images generated from noise by GAN and WGAN-GP with the same architecture. D is a CNN with 4 convolutional layers. G is a deconvolution network that takes in a 128-element-noise vector, processes through 4 deconvolutional layers and output a $64\times 64$ image. }
    \label{fig:mode-collapse}
\end{figure}

Figure \ref{fig:other-output}(e) and (k) show strong mode collapse with several images being clearly identical, consistent with existing literature (Goodfellow, 2016 and reference therein). Normally we would not expect to observe mode collapse in super-resolution images, since they are enforced to look similar to input files by the $L_1$ loss term in the generator loss (see Section \ref{section:perceptual-loss}). Their appearance suggests that in these two examples, the $L_1$ loss term should have a larger weight. 

To further investigate if mode collapse appears, we use two DCGANs to generate face images from 128-element random noise vectors. One of the DCGANs uses the GAN objective and the other use the WGAN-GP objective. Figure \ref{fig:mode-collapse} shows non-cherry-picked images generated by the two DCGANs. Neither outputs show significant mode collapse. DCGAN with the GAN objective does generate a few images that clearly share features. For example images in (row 1, column 7), (row 2 column 7), (row 4 column 5) show very similar smiles. In comparison, DCGAN with WGAN-GP seem to have generated more diverse faces.

\section{Conclusion}
\label{gen_inst}
We reviewed the implementation difference between GAN, WGAN and WGAN-GP objectives and combined the above training objectives with MLP, CNN and ResNet architectures to evaluate the effectiveness of Wasserstein GAN for single image super resolution. Our results verify that the Wasserstein GAN objective facilitates stable GAN training on a wide range of architectures, and that the discriminator loss / Wasserstein distance provides a metric that correlates well with training progress. 

\section*{References}
\medskip
\small
Arjovsky, M.,\ \& Bottou, L. (2017a). Towards principled methods for training generative adversarial networks. In {\it NIPS 2016 Workshop on Adversarial Training}. 

Arjovsky, M., Chintala, S.,\ \& Bottou, L. (2017b). Wasserstein gan. {\it arXiv preprint arXiv:1701.07875}.

Finn, C., Christiano, P., Abbeel, P.,\ \& Levine, S. (2016). A connection between generative adversarial networks, inverse reinforcement learning, and energy-based models. {\it arXiv preprint arXiv:1611.03852}.

Goodfellow, Ian J. (2014) On distinguishability criteria for estimating generative models. {\it arXiv preprint arXiv:1412.6515}.

Goodfellow, I. (2016). NIPS 2016 Tutorial: Generative Adversarial Networks. {\it arXiv preprint arXiv:1701.00160}.

Gulrajani, I., Ahmed, F., Arjovsky, M., Dumoulin, V.,\ \& Courville, A. (2017). Improved Training of Wasserstein GANs. {\it arXiv preprint arXiv:1704.00028}.

Liu, Z., Luo, P., Wang, X.,\ \& Tang, X. (2015). Deep learning face attributes in the wild. In {\it Proceedings of the IEEE International Conference on Computer Vision} (pp. 3730-3738).

Isola, P., Zhu, J. Y., Zhou, T.,\ \&  Efros, A. A. (2016). Image-to-image transla- tion with conditional adversarial networks. {\it arXiv preprint arXiv:1611.07004}.

Reed, S., Akata, Z., Yan, X., Logeswaran, L., Schiele, B.,\ \& Lee, H. (2016). Generative adversarial text to image synthesis. {\it arXiv preprint arXiv:1605.05396}.

Zhang, H., Xu, T., Li, H., Zhang, S., Huang, X., Wang, X.,\ \& Metaxas, D. (2016). Stackgan: Text to photo-realistic image synthesis with stacked generative adversarial networks. {\it arXiv preprint arXiv:1612.03242}.

Zhu, J. Y., Kr{\"a}henb{\"u}hl, P., Shechtman, E.,\ \& Efros, A. A. (2016). Generative visual manipulation on the natural image manifold. In {\it European Conference on Computer Vision}, pages 597–613. Springer.

\end{document}